\documentclass[10pt, a4paper]{article}
\usepackage{lrec2022} % this is the new LREC2022 Style
\usepackage{multibib}
\newcites{languageresource}{Language Resources}
\usepackage{graphicx}
\usepackage{tabularx}
\usepackage{soul}
\usepackage{titlesec}
\titleformat{\section}{\normalfont\large\bfseries\center}{\thesection.}{1em}{}
\titleformat{\subsection}{\normalfont\SmallTitleFont\bfseries\raggedright}{\thesubsection.}{1em}{}
\titleformat{\subsubsection}{\normalfont\normalsize\bfseries\raggedright}{\thesubsubsection.}{1em}{}
\renewcommand\thesection{\arabic{section}}
\renewcommand\thesubsection{\thesection.\arabic{subsection}}
\renewcommand\thesubsubsection{\thesubsection.\arabic{subsubsection}}
\usepackage{epstopdf}
\usepackage[utf8]{inputenc}
\usepackage{color}
\usepackage{multirow}
\usepackage{hyperref}
\usepackage{xstring}
\usepackage{tablefootnote}
\usepackage{hyperref}
\usepackage[table,dvipsnames]{xcolor}

\usepackage{enumitem}
\setlist[enumerate]{noitemsep, nolistsep}
\setlist[itemize]{noitemsep, nolistsep}

\usepackage{gb4e}
\noautomath

\title{CrudeOilNews: An Annotated Crude Oil News Corpus for Event Extraction}

\name{Meisin Lee, Lay-Ki Soon, Eu-Gene Siew, Ly Fie Sugianto} 

\address{Monash University \\
         Malaysia Campus, Jalan Lagoon Selatan Bandar Sunway, 47500 Selangor, Malaysia \\
         \{mei.lee, soon.layki, siew.eu-gene, lyfie.sugianto\}@monash.edu\\}

\abstract{
In this paper, we present CrudeOilNews, a corpus of English Crude Oil news for event extraction. It is the first of its kind for Commodity News and serve to contribute towards resource building for economic and financial text mining. This paper describes the data collection process, the annotation methodology and the event typology used in producing the corpus. Firstly, a seed set of 175 news articles were manually annotated, of which a subset of 25 news were used as the adjudicated reference test set for inter-annotator and system evaluation. Agreement was generally substantial and annotator performance was adequate, indicating that the annotation scheme produces consistent event annotations of high quality. Subsequently the dataset is expanded through (1) data augmentation and (2) Human-in-the-loop active learning.  The resulting corpus has 425 news articles with approximately 11k events annotated. As part of active learning process, the corpus was used to train basic event extraction models for machine labeling, the resulting models also serve as a validation or as a pilot study demonstrating the use of the corpus in machine learning purposes. The annotated corpus is made available for academic research purpose \href{https://github.com/meisin/CrudeOilNews-Corpus}{https://github.com/meisin/CrudeOilNews-Corpus}. \\ \newline  }

\begin{document}

\maketitleabstract

\section{Introduction}

%Commodity news contain a wealth of information; one of the key information is the analysis of recent commodity price movements along with notable events that led to that movement. Through event extraction, useful information can be extracted from commodity news. However there is a lack of annotated dataset for commodity news. To facilitate future research in this area, w
Financial markets are sensitive to breaking news on economic events. Specifically for crude oil markets, it is observed in \cite{brandt2019macro} that news about macroeconomic fundamentals and geopolitical events affect the price of the commodity. Apart from fundamental market factors, such as supply, demand, and inventory, oil price fluctuation is strongly influenced by economic development, conflicts, wars, and breaking news \cite{wu2021effective}. Therefore, accurate and timely automatic identification of events in news items is crucial for making timely trading decisions. 
%Unstructured data originating from news articles can to be mined in order to extract knowledge useful for guiding decision making processes. %The analyses by the reporter or journalist of past events provide a good distillation of world events that are truly causal to the movement of commodity prices. 
Commodity news typically contain these few key information: (i) analysis of recent commodity price movements (up, down or flat), (ii) a retrospective view of notable event(s) that led to such a movement, and (iii) forecast or forward-looking analysis of supply-demand situation as well as projected commodity price targets. 
Here is a snippet taken from a piece of crude oil news:
\begin{exe}
		\ex U.S. crude stockpiles \textbf{soared} by 1.350 million barrels in December from a mere 200 million barrels to 438.9 million barrels, due to this \textbf{oversupply} crude oil prices \textbf{plunged} more than 50\% as Tuesday.
\end{exe}

There is a small number of corpora in the Finance and Economics domain such as SENTiVENT in \cite{jacobs2021sentivent}, but all are focused on company-specific events, and are used mainly for the purpose of stock price prediction. As acknowledged by \cite{jacobs2021sentivent}, due to the lack of annotated dataset in the Finance and Economics domain, only a handful of supervised approaches exist for Financial Information Extraction. To the best of our knowledge, there is no available annotated corpus for crude oil or any other commodities. We aim to contribute towards resource building and facilitate future research in this area by introducing CrudeOilNews corpus, an event extraction dataset focusing on Macro-economic, Geo-political, and crude oil supply and demand events. The annotation schema used are aligned to ACE (Automatic Content Extraction) and ERE (Entities, Relations, and Events) standards, so event extraction systems developed for ACE/ERE can be used readily on this corpus.

The contributions of this work are as follows:
\begin{itemize}
    \item Introduced CrudeOilNews corpus, the first annotated corpus for crude oil consisting of 425 crude oil news articles. It is a ACE/ERE-like corpus with the following annotated: (i) Entity mentions, (ii) Events (triggers and argument roles), and (iii) Event Properties (Polarity, Modality, and Intensity;
    \item Introduced a new event property to capture a more complete representation of events. The new property -\texttt{INTENSITY} captures the state of an existing event whether it further intensifies or eased;
    \item Addressed the obvious class-imbalance in event properties by over-sampling minority classes and adding them into corpus through data augmentation;
    \item Used Human-in-the-Loop Active Learning to expand the corpus with model inference while optimizing human annotation effort to focus on just less confident (and likely less accurate) predictions. 
\end{itemize}

%Specifically for this genre of text, it is insufficient to conduct Event Detection without extracting event arguments as well. To illustrate this point, consider event E3 in example sentence (1), trigger word "plunged", it is incomplete to just identify the event \textbf{movement-down-loss} without knowing "what" \textit{plunged}. In commodity news, possible events are \textit{`price plunged'}, \textit{`supply plunged'} and \textit{`demand plunged'}. As seen sentence (1), the sentence consists of two movement events (E1 and E3). To disambiguate and to extract events accurately, it is important to extract event arguments accurately as well.

%In this paper, information about the corpus, including data collection and annotation procedure are laid out in Section \ref{sec:Corpus}. Annotation methodologies for \textbf{Entities} and \textbf{Events} (Event Trigger, Event Arguments, and Event Metadata are covered in detail in Section \ref{sec:Annotation}. Annotation evaluation and challenges are stated in Section \ref{sec:challenges}. Potential uses of the corpus are covered in Section \ref{sec:Uses}; it is then followed by Conclusion in Section \ref{sec:Conclusion}.

\section{Related Work}
\subsection{Annotation Methodologies}
The annotation methodologies presented here are conceived based on the annotation standards of ACE, and ERE. An extensive comparison has been made in \cite{aguilar2014comparison}, where authors analyzed and provided a summary of the different annotation approaches. Subsequently there was a number of works that expanded earlier annotation standards, such as in \cite{ogorman-etal-2016-richer}, authors introduced the Richer Event Description (RED) corpus and methodologies that annotate entities, events, times, entities relations (co-reference and partial co-reference), and events relations (temporal, causal, and sub-events). We have strived to align to ACE/ERE programs as closely as possible, but have made minor adaptations to cater to special characteristics found in crude oil news. \textit{Tense} and \textit{Genericity} defined in ACE2005 are dropped from our annotation scope while the new property - \textit{Intensity} is introduced.

\subsection{Finance and Economic Domain} \label{subsec:financeResources}
In the domain of Finance and Economics, majority of the available datasets are on \textbf{company-related events} and are used mainly for extracting company-related events for company stock price prediction. A variety of methods are used in economic event detection, such as hand-crafted rule-sets, ontology knowledge-bases, and using techniques like distant, weakly or semi-supervised training. For a more targeted discussion, we focus only on manually annotated datasets suitable for \textbf{supervised} training. 

In \cite{jacobs2018economic}, \cite{lefever2016classification}, authors introduced a dataset focused on annotating continuous trigger spans of 10 types and 64 subtypes of company-economic events in a corpus of English and Dutch economic text, some examples of event types are \textit{Buy ratings, Debt, Dividend, Merger \& acquisition, Profit, Quarterly results}. As a continuation of the work, the authors introduced SENTiEVENT, a fine-grained ACE/ERE-like dataset in \cite{jacobs2021sentivent}. Just like the earlier work, their focus is mainly on company-related and financial events. Among the list of defined event topology, the only category that has an overlap to our work is ``Macroeconomics'', an event category that captures a broad range of events that is not company-specific such as economy-wide phenomena, and governmental policy in news. While they choose to remain at a broad level, our work compliments theirs by defining key events in the Macro-economic and Geo-political category at a detailed-level.

As part of the search for commodity news related resources, we came across RavenPack's\footnote{RavenPack is an analytics provider for financial services. Among their services are finance and economic news sentiment analysis. More information can be found on their page: https://www.ravenpack.com/} crude oil dataset. This dataset is available through subscription at the Wharton Research Data Services (WRDS). It is made up of news headlines and a corresponding sentiment score generated by Ravenpack's own analytic engine. Unfortunately this dataset is not suitable for the task of \textit{supervised} event extraction as it only contains sentiment score without any event annotations. However, Ravenpack's event taxonomy on crude oil-related events proves to be a useful resource in helping us define our own event typology. Details of event typology is covered in Section \ref{subsection:eventTaxanomy}.

\section{Dataset Collection}
First, we crawled crude oil news articles from \textit{investing.com}\footnote{\href{https://www.investing.com/commodities/crude-oil-news}{https://www.investing.com/commodities/crude-oil-news}}, a financial platform and financial/business news aggregator and is considered one of the top three global financial websites in the world.
%\footnote{Due to copyright issues, original articles are not released. Instead URL links to each articles are provided along with their corresponding annotation file.}  - news aggregator.

We crawled news articles dating from Dec 2015 to Jan 2020 (50 months). From the pool of crude oil news, we uniformly sampled 175 pieces of news articles throughout the 50-month period to ensure events are evenly represented and not skewed towards a certain topic of a particular time window. These 175 news articles were duly annotated by two annotators and they form the gold-standard annotation. For the purposes of assessing the inter-annotator agreement and evaluating the annotation guidelines, 25 news were selected out of the gold-standard dataset as the adjudicated set (ADJ).

\begin{figure*}[htbp]
\begin{center}
%\fbox{\parbox{6cm}{
%This is a figure with a caption.}}
% old picture \includegraphics[scale=0.5]{lrec2020W-image1.eps} 
\includegraphics[scale=0.5]{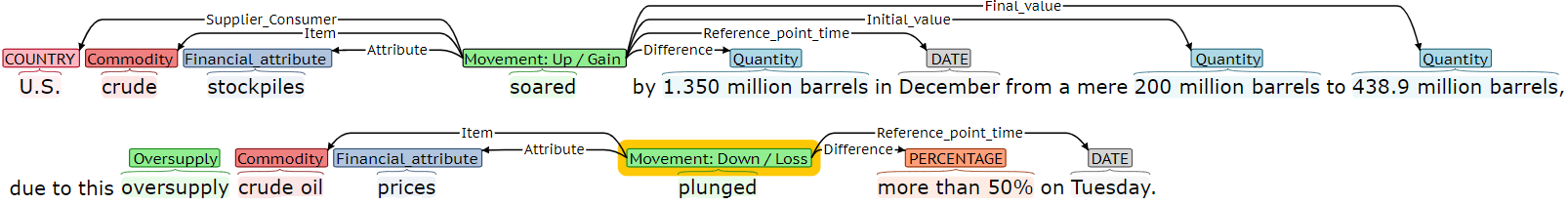} 
\caption{Annotation using Brat annotation tool: (i) entities (both nominal and named) are annotated with entity types listed above the respective words (in various colours except green), (ii) events trigger words are also annotated (in green), and (iii) entities are linked to their respective event trigger through arches, argument roles these entities play in linked events are listed on the arches.
Note: Event properties: modality, polarity and intensity are not shown in this diagram} 
\vspace{-0.7em}
\label{fig:annotation}
\end{center}
\end{figure*}

\section{Annotation Setup}  \label{subsec:annotation_setup}
The dataset is annotated using \textbf{Brat rapid annotation tool} \cite{stenetorp2012brat}, a web-based tool for text annotation. An example of a sentence annotated using Brat is shown in Figure \ref{fig:annotation}. 

The annotation process is designed to have high inter-annotator agreement (IAA). One of the criteria is that the annotators should possess domain knowledge in business, finance, and economics. It is imperative for the annotators to understand financial and macro-economic terms and concepts to interpret the text accurately and annotate events accordingly. For instance, sentences containing macro-economic terms such as \textit{contango}, \textit{quantitative easing}, and \textit{backwardation} will require annotators to have finance and economics domain knowledge. To meet this criteria, we recruited two annotators from a pool of undergraduate students from the School of Business of a local university. Annotators were then given annotation training and provided with clear annotation schemas and examples. Every piece of text was duly annotated by two annotators independently. 

The annotation was done based on the following layers sentence by sentence:
\begin{itemize}
    \item Layer 1: Identify and annotate entity mentions.
    \item Layer 2: Annotate events by identifying event triggers.
    \item Layer 3: Using event triggers as anchors, identify and link surrounding entity mentions to their respective events. Annotate the argument roles each entity mention plays with respect to the events identified. 
    \item Layer 4: Annotate event properties: modality, polarity and intensity.
\end{itemize}

After each layer, an adjudicator assessed the annotation and evaluated inter-annotator agreement before finalizing the annotation. For cases where there are annotation discrepancies, the adjudicator will act as the tie-breaker to decide on the final annotation. Once finalized, annotators then proceed with the next layer. This is done to ensure no accumulation of the previous layer's errors in the subsequent layers of annotation.

\subsection{Annotation Guidelines}
This section describes our definition of events and principles for annotation of entity mentions and events. Annotation of events is further divided into (i) annotating entity mentions, (ii) annotating event triggers, (iii) linking entity mentions to their respective events and identifying the argument roles each entity plays, (iv) assigning the right property labels to \textit{Polarity}, \textit{Modality} and \textit{Intensity}.

\subsection{Entity Mention}
An entity mention is a reference to an object or a set of objects in the world, including named entities, nominal entities, and pronouns. For simplicity and convenience, \textbf{values} and \textbf{temporal expressions} are also considered as entity mentions in this work. There are 21 entity types identified and annotated in the dataset, see Appendix \ref{app:entity} for the full list. Nominal entities relating to Finance and Economics are annotated. Apart from crude oil-related terms, below here are some examples of nominal entities found in the corpus and was duly annotated:
\begin{itemize}
    \item attributes : \textit{price, futures, contract, imports, exports, consumption, inventory, supply, production}
    \item economic entity : \textit{economic growth, economy, market(s), economic outlook, growth, dollar}
\end{itemize}
%The list of entity types and the corresponding list of entities (both nominal and named entities) are listed in Appendix , the list of entity types are listed in Appendix \ref{appendix:entityTypes}Se\ref{appendix:entityTypes}.

\subsection{Events}
Events are defined as `specific occurrences', involving `specific participants'.
%\footnote{\href{https://www.ldc.upenn.edu/sites/www.ldc.upenn.edu/files/english-events-guidelines-v5.4.3.pdf}{https://www.ldc.upenn.edu/sites/www.ldc.upenn.edu/files/english-events-guidelines-v5.4.3.pdf}}. 
The occurrence of an event is marked by the presence of an event trigger. In addition to identifying triggers, all of the participants of each event are also identified. An event's participants are \textit{entities} that play a role in that Event. Details and rules for identifying event triggers and event Arguments are covered below:

\paragraph{Event Triggers} \label{para:event_triggers} Our annotation of \textit{event trigger} is aligned to ERE where an event trigger (known as event nugget in the shared task in \cite{mitamura2015event}) can be either a single word  (main verb, noun, adjective, adverb) or a  continuous multi-word phrase. 
%For our work, we have excluded discontinuous trigger spans from our scope. 
Here are some examples found in the dataset: 
\begin{itemize}
    \item \textbf{Verb}: Houti rebels \textbf{attacked} Saudi Arabia.
    \item \textbf{Noun}: The government slapped \textbf{sanctions} against its petroleum....
    \item \textbf{Adjective}: A fast \textbf{growing} economy has...
    \item \textbf{Multi-verb}: The market \textbf{bounced back}....
\end{itemize}

\textit{Event trigger} is the minimal span of text that most succinctly expresses the occurrence of an event. Annotators are instructed to keep the trigger as small as possible while maintaining the core lexical semantics of the event. For example, for phrase ``Oil price \textbf{edged lower}'', only the trigger word ``lower'' is annotated. 

\paragraph{Event Arguments} \label{subsection:eventArguments}
After event triggers and entity mentions are annotated, entities need to be linked up to form events. An event contains an event trigger and a set of event arguments. 
%Details of events and their corresponding list of arguments are found in Appendix \ref{appendixEventArguments}. 
Referring to Figure~\ref{fig:annotation}, the event trigger \textbf{soared} is linked to seven entity mentions via arches. The argument role of each entity mention is labeled on each arch respectively, while entity types are labeled in various colours on top of each entity span. This information is also summarized in tabular format in Table~\ref{table:EventArguments}. 

\vspace{-0.5em}
\begin{table}[h!]   
    \centering \small        
    \caption{List of Event Arguments of example shown in Figure \ref{fig:annotation}}.
    \begin{tabular}{ l | l}  \hline
    \textbf{Entity} & \textbf{Argument Role} \\ \hline
    U.S. & {\fontfamily{qcr}\selectfont SUPPLIER} \\ \hline
    crude & {\fontfamily{qcr}\selectfont ITEM} \\ \hline
    stockpiles & {\fontfamily{qcr}\selectfont ATTRIBUTE}\\ \hline
    1,350 million barrels & {\fontfamily{qcr}\selectfont DIFFERENCE} \\ \hline
    December & {\fontfamily{qcr}\selectfont REFERENCE\_POINT\_TIME} \\ \hline
    200 million barrels & {\fontfamily{qcr}\selectfont INITIAL\_VALUE} \\ \hline
    438.9 million barrels & {\fontfamily{qcr}\selectfont FINAL\_VALUE}  \\ \hline
    \end{tabular}
    \label{table:EventArguments}
\end{table}
\vspace{-0.7em}

\subsubsection{Event Typology} \label{subsection:eventTaxanomy}
According to \cite{brandt2019macro} who analyzed Ravenpack's sentiment score of each event type and oil price, events that move commodity prices are \textbf{geo-political}, \textbf{macro-economic} and \textbf{commodity supply and demand} in nature. Based on Ravenpack's event taxonomy, we have defined a set of 18 oil-related event types as our annotation scope. Event types and the corresponding list of example trigger words and example key arguments are listed out in Table \ref{table:EventTypes}. See Appendix \ref{app:schema} for event schema for all 18 event types.

 \begin{table*}[h!]   
    \centering \small
    \caption{List of Event types with example trigger words and example key arguments.}
    \begin{tabular}{ p{4.2cm} | p{5.5cm} | p{5cm}}  \hline
    \textbf{Event Type} & \textbf{Example Trigger Word(s)} & \textbf{Example key arguments} \\ \hline
    1. {\fontfamily{qcr}\selectfont CAUSED-MOVEMENT-DOWN-LOSS} & \textit{cut, trim, reduce, disrupt, curb, squeeze, choked off} & oil production, oil supplies, interest rate, growth forecast \\ \hline
    2. {\fontfamily{qcr}\selectfont CAUSED-MOVEMENT-UP-GAIN} & \textit{boost, revive, ramp up, prop up, raise} & oil production, oil supplies, growth forecast\\ \hline
    3. {\fontfamily{qcr}\selectfont CIVIL-UNREST} & \textit{violence, turmoil, fighting, civil war, conflicts} & Libya, Iraq \\ \hline
    4. {\fontfamily{qcr}\selectfont CRISIS} & \textit{crisis, crises} & debt, financial\\ \hline
    5. {\fontfamily{qcr}\selectfont EMBARGO} & \textit{embargo, sanction} & Iraq, Russia  \\ \hline
    6. {\fontfamily{qcr}\selectfont GEOPOLITICAL-TENSION} & \textit{war, tensions, deteriorating relationship} & Iraq-Iran\\ \hline
    7. {\fontfamily{qcr}\selectfont GROW-STRONG} & \textit{grow, picking up, boom, recover, expand, strong, rosy, improve, solid} & oil production, economic growth, U.S. dollar, crude oil demand \\ \hline
    8. {\fontfamily{qcr}\selectfont MOVEMENT-DOWN-LOSS} & \textit{fell, down, less, drop, tumble, collapse, plunge, downturn, slump, slide, decline} & crude oil price, U.S. dollar, gross domestic product (GDP) growth\\ \hline
    9. {\fontfamily{qcr}\selectfont MOVEMENT-FLAT} & \textit{unchanged, flat, hold, maintained} & oil price \\ \hline
    10. {\fontfamily{qcr}\selectfont MOVEMENT-UP-GAIN} & \textit{up, gain, rise, surge, soar, swell, increase, rebound} & oil price, U.S. employment data, gross domestic product (GDP) growth \\ \hline
    11. {\fontfamily{qcr}\selectfont NEGATIVE-SENTIMENT} & \textit{worries, concern, fears} \\ \hline
    12. {\fontfamily{qcr}\selectfont OVERSUPPLY} & \textit{glut, bulging stock level, excess supplies} \\ \hline
    13. {\fontfamily{qcr}\selectfont POSITION-HIGH} & \textit{high, highest, peak, highs} \\ \hline
    14. {\fontfamily{qcr}\selectfont POSITION-LOW} & \textit{low, lowest, lows, trough} \\ \hline
    15. {\fontfamily{qcr}\selectfont PROHIBITION} & \textit{ban, bar, prohibit} & exports, imports \\ \hline
    16. {\fontfamily{qcr}\selectfont SHORTAGE} & \textit{shortfall, shortage, under-supplied} & oil supply\\ \hline
    17. {\fontfamily{qcr}\selectfont SLOW-WEAK} & \textit{slow, weak, tight, lackluster, falter, weaken, bearish, slowdown, crumbles} & global economy, regional economy, economic outlook, crude oil demand \\ \hline
    18. {\fontfamily{qcr}\selectfont TRADE-TENSIONS} & \textit{price war, trade war, trade dispute} & U.S.-China \\ \hline
    \end{tabular}
    \vspace{-1.0em}
    \label{table:EventTypes}
\end{table*}

\subsubsection{Event Property: Event Polarity, Modality and Intensity} \label{subsection:metadata}
After events are identified, they are also assigned a label each for the properties respectively. 

\paragraph{POLARITY} ({\fontfamily{qcr}\selectfont POSITIVE} and {\fontfamily{qcr}\selectfont NEGATIVE}) \\ An event has the value {\fontfamily{qcr}\selectfont POSITIVE} unless there is an explicit indication that the event did not take place, in which case {\fontfamily{qcr}\selectfont NEGATIVE} is assigned.

%\begin{exe}
%    \item OPEC countries \textcolor{red}{\textit{refused}} to \textbf{cut} oil supplies [\textcolor{red}{\fontfamily{qcr}\selectfont NEGATIVE}]
%\end{exe}

%Currently the ending of an event is annotated as Negative. eg: Put an end to the crisis, defy, break

\paragraph{MODALITY} ({\fontfamily{qcr}\selectfont ASSERTED} and {\fontfamily{qcr}\selectfont OTHER}) \\ Event modality determines whether the event represents a ``real'' occurrence. {\fontfamily{qcr}\selectfont ASSERTED} is assigned if the author or speaker refers  to it as though it were a real occurrence, and {\fontfamily{qcr}\selectfont OTHER} otherwise. {\fontfamily{qcr}\selectfont OTHER} covers believed events, hypothetical events, commanded and requested event, threats, proposed events, discussed events, desired events, promised events, and other unclear construct.

%\begin{exe}
%    \item The market \textcolor{blue}{\textit{expects}} US to \textbf{sanction} Iran. [\textcolor{blue}{\fontfamily{qcr}\selectfont OTHER}]
%\end{exe}

\paragraph{INTENSITY} ({\fontfamily{qcr}\selectfont NEUTRAL}, {\fontfamily{qcr}\selectfont INTENSIFIED}, and {\fontfamily{qcr}\selectfont EASED}) \\
Event intensity is a new event property, specifically created for this work to better represent events found in this corpus. Oftentimes, events reported in Crude Oil News are about the intensity of an existing event, whether the event is further intensified or eased. 

Examples of events where one is {\fontfamily{qcr}\selectfont INTENSIFIED} and the other one {\fontfamily{qcr}\selectfont EASED}:
\vspace{-0.5em}
\begin{exe}
    \item ...could hit Iraq 's output and \textcolor{ForestGreen}{\textit{deepen}} a supply \textbf{shortfall}. [\textcolor{ForestGreen}{\fontfamily{qcr}\selectfont INTENSIFIED}]
    \item Libya 's civil \textbf{strife} has been \textcolor{OliveGreen}{\textit{eased}} by potential peace talks. [\textcolor{Green}{\fontfamily{qcr}\selectfont EASED}]
\end{exe}
\vspace{-0.5em}
The event \textbf{strife} (civil unrest) in sentence (2) is not an event with negative polarity because the event has actually taken place but with reduced intensity. INTENSITY label is used to capture the interpretation accurately, showing that the civil unrest event has indeed taken place but now with updated `intensity'. 

With these three event properties, we can annotate and capture all essential information about an event. To further illustrate this point, consider the list of examples of complex events below:

\vspace{-0.5em}
\begin{exe}
    \item OPEC \textcolor{red}{\textit{cancelled}} a \textcolor{blue}{\textit{planned}} \textcolor{ForestGreen}{\textit{easing}} of output \textbf{cuts}. [\textcolor{red}{\fontfamily{qcr}\selectfont NEGATIVE}, \textcolor{blue}{\fontfamily{qcr}\selectfont OTHER}, \textcolor{ForestGreen}{\fontfamily{qcr}\selectfont EASED}] 
    \item In order to end the global crisis, OPEC may \textcolor{red}{\textit{hesitate}} to implement a \textcolor{blue}{\textit{planned}} \textcolor{ForestGreen}{\textit{loosening}} of output \textbf{curbs}. [\textcolor{red}{\fontfamily{qcr}\selectfont NEGATIVE}, \textcolor{blue}{\fontfamily{qcr}\selectfont OTHER},
    \textcolor{ForestGreen}{\fontfamily{qcr}\selectfont EASED}]
    
    \item Oil prices rose to \$110 a barrel on \textcolor{blue}{\textit{rumours}} of a \textcolor{ForestGreen}{\textit{renewed}} \textbf{strife}. [\textcolor{red}{\fontfamily{qcr}\selectfont POSITIVE}, \textcolor{blue}{\fontfamily{qcr}\selectfont OTHER},
    \textcolor{ForestGreen}{\fontfamily{qcr}\selectfont INTENSIFIED}]
\end{exe}
\vspace{-0.5em}

%\subsubsection{Example of a sub-subsection with a long heading that will occupy two lines}
%
%Yet another example of a sub-subsection. Yet another example of a sub-subsection. Yet another example of a sub-subsection. Yet another example of a sub-subsection. Yet another example of a sub-subsection.

\subsection{Inter-Annotator Agreement}
%%% https://towardsdatascience.com/inter-rater-agreement-kappas-69cd8b91ff75
%%% https://towardsdatascience.com/inter-annotator-agreement-2f46c6d37bf3
Inter-annotator agreement (IAA) is a good indicator of how clear our annotation guidelines are, how uniformly annotators understand it, how robust are the event typology and overall how feasible the annotation task is. We evaluate IAA on each annotated category separately (see Table \ref{table:IAA} for the list) using the most commonly measurement: Cohen's Kappa, with the exception of \textit{entity spans} and \textit{trigger spans}. These two annotations are made at token level, forming spans of a single token or multiple continuous tokens. For the sub-tasks of \textit{entity mention detection} and \textit{trigger detection}, the token-level span annotation were unitized to compute IAA, this approach is similar to unitizing and measuring agreement in Named Entity Recognition\cite{mathet2015unified}. According to \cite{hripcsak2005agreement}, Cohen's kappa is not the most appropriate measurement for IAA in Named Entity Recognition. In \cite{deleger2012building}, authors provided an in-depth analysis of why is the case and proposed the use of pairwise F1 score as the measurement. Hence for the evaluation of \textit{entity spans} and \textit{trigger spans}, we report on both F1 as well as “token-level” kappa. Both score were measured without taking into account the un-annotated tokens - labelled "O". 

As for the rest of the annotation category, we report only on Cohen's Kappa as this is the standard measure of IAA for classification task. We calculate the agreement by comparing annotation outcomes of the two annotators with each other, arbitrarily treating one as the `gold' reference. We also scored each annotator separately on the adjudicated (ADJ) set. The ADJ set consists of 25 documents collected through correcting and combining the manual annotations of these documents by the adjudicator. The final scores are calculated by averaging the results across all comparisons. Table \ref{table:IAA} shows the average agreement scores for all annotation categories. 

Event nugget scoring method introduced in \cite{liu2015evaluation} was not used here because their assessment is rolled up into ``Span'', ``Type'', and ``Realis'', too coarse to show IAA on each annotation category.

\begin{table}[h!]   
    \centering \small
    \caption{Inter-Annotator Agreement (IAA) for all annotation categories. For categories involving spans (marked by$^{*}$), both Cohen's kappa (calculated on ``token level'') and F1 score measurements are provided.}
    \begin{tabular}{ l | c | c }  \hline
    \textbf{Task} & \textbf{Cohen's Kappa $\kappa$ } & \textbf{F1 Score} \\ \hline
    Entity spans$^{*}$ & 0.82 & 0.91 \\ \hline
    Trigger spans$^{*}$ & 0.68 & 0.75 \\ \hline
    Entity Type & 0.89 & - \\ \hline
    Event Type & 0.79 & - \\ \hline
    Argument Role & 0.78 & - \\ \hline
    Event Polarity & 0.70 & - \\ \hline
    Event Modality & 0.63 & - \\ \hline
    Event Intensity & 0.59 & - \\ \hline
    \end{tabular}
    \label{table:IAA}
    \vspace{-0.7em}
\end{table}

\paragraph{Analysis}
We benchmark these IAA scores with the `strength of agreement' of each Kappa ranges as set out by \cite{landis1977measurement}. Most annotation categories achieved \textit{substantial agreement} with the exception of \textit{Intensity} classification. This is because classifying \textit{Intensity} is more challenging where some of the cue words for determining the event intensity are themselves trigger words. For example:
\begin{exe}
    \item \textbf{Oversupply} could \underline{rise} next year when Iraq starts to export more oil. 
\end{exe}

The word \textit{rise} here is a cue word to indicate that oversupply might be further INTENSIFIED but it also could be misinterpreted as another separate event. On the other hand, we achieve very high agreement on identifying entity spans. This is because entities in the news articles are majority Named Entities with very clear span boundaries, and classifying the entities to the correct entity type is also rather straight forward. Even for nominal entities such as \textit{crude oil, oil markets}, and etc, their span boundaries are clear.
%The main area of  of event annotation is ambiguities on `eventiveness'. Here are examples where information is conveyed in an indirect way that makes it difficult to pinpoint any clear-cut events:
%\vspace{-0.5em}
%        \begin{exe}
%            \item Then spent the rest of the week trying to defend those gains as market optimism ......
 %           \item  Oil prices felt pressure on Tuesday from news that....
%          \end{exe}
%        \vspace{-0.5em}
%In cases above where an event trigger is absent, we do not annotate any event even though the event is conveyed in an implicit manner. Other challenges include knowing idiomatic expressions in order to accurately annotate event nuggets. Because such expressions are often cultural, it could be an challenge to arrive at an consensus on idiomatically expressed events such as ’kicked the bucket’.

The common mistake in trigger span detection and classification is the different interpretation of the minimum span of an event trigger. Examples of common annotation errors are: (i) the trigger word for ``crude oil \textbf{inched higher}" should be just ``higher'', and (ii) "Oil \textbf{pursued an upward trend}" should be just ``upward trend''.

From the cases where annotators disagree, we analyze and found that most of them stem from differences in interpreting special concepts for example: 
\begin{itemize}
    \item The word \textit{outlook}, should it be interpreted as \textit{forecast}? Or, should it be considered as a cue word for event modality?
    \item If events surrounding US \textit{employment} data are annotated, then what about \textit{unemployment}? Should this be treated as employment data but negated using negative polarity? 
    \item How should double negation be treated?  For example, `failed attempt to prevent a steep drop in oil prices', both \textit{failed} and \textit{prevent} are considered negative polarity cue words, creating a double negation situation.
    %% fear and worry - should it be negative sentiment or polarity cue word?
\end{itemize}
For these non-straight forward cases, each one was handled on a case-by-case basis where the adjudicator discussed each situation with the annotators to seek consensus before finalizing an agreed annotation.

%Different interpretation in classifying event types, eg: uncertain: is it negative sentiment or slow-weak?

\section{Expanding the Dataset}
Manual annotations are labour-intensive and time-consuming, as this is seen in our gold-standard manual annotation where it consists only 175 documents or news articles. In order to produce a sufficiently large dataset useful for supervised event extraction, we utilize (1) Data Augmentation and (2) Human-in-the-Loop Active Learning.

\begin{table*}[h!]   
    \centering \small
    \caption{Event Properties Distribution and classification results (F1-score) before and after data augmentation. Prior to data augmentation: the corpus shows obvious class imbalance for all three event properties. Post-Data Augmentation: Class distribution is slightly adjusted and F1-scores for minority classes improved accordingly.} 
    \begin{tabular}{ l |r | c | c | c | c | c |c }  \hline
    & \multicolumn{2}{c|}{\textbf{Gold Dev Set}} &  \textbf{Before} & \textbf{Augmentation} & \multicolumn{2}{c |}{\textbf{Updated Count}} & \textbf{After} \\ \hline
    \textbf{Event Properties} & \textbf{Ratio} & \textbf{\# Events} & \textbf{F1} & \textbf{\# Events} & \textbf{Ratio} & \textbf{\# Events}\ & \textbf{F1} \\ \hline
    Polarity: {\fontfamily{qcr}\selectfont POSITIVE} & 97.01\% & 2,855 & 0.76 & 965 & 95.40\% & 3,820 & 0.76\\ \hline
    Polarity: {\fontfamily{qcr}\selectfont NEGATIVE} & 2.99\% & 88 & \cellcolor{Lavender}{0.24} & 96 & 4.60\% & 184 & \cellcolor{YellowGreen}{0.39}\\ \hline \hline
    Modality: {\fontfamily{qcr}\selectfont ASSERTED} & 82.94\% & 2,441 & 0.71 &  771 & 80.22\% & 3,212 & 0.74 \\ \hline
    Modality: {\fontfamily{qcr}\selectfont OTHER} & 17.06\% & 502 & \cellcolor{Lavender}{0.35} & 290 & 19.78\% & 792 & \cellcolor{YellowGreen}{0.42} \\ \hline \hline
    Intensity: {\fontfamily{qcr}\selectfont NEUTRAL} & 93.78\% & 2,760 & 0.76 &  745 & 87.54\% & 3,505 & 0.85 \\ \hline
    Intensity: {\fontfamily{qcr}\selectfont EASED} & 3.64\% & 107 & \cellcolor{Lavender}{0.36} & 196 & 7.57\% & 303 & \cellcolor{YellowGreen}{0.49} \\ \hline
    Intensity: {\fontfamily{qcr}\selectfont INTENSIFIED} & 2.58\% & 77 & \cellcolor{Lavender}{0.25} & 120 & 4.90\% & 196 & \cellcolor{YellowGreen}{0.37} \\ \hline 
    \end{tabular}
    \label{table:EventPropertyDistribution}
\end{table*}

\subsection{Data Augmentation} \label{sec:data_augmentation}
The main purpose of introducing augmented data is to address the issue of serious class imbalance in Event Properties in the dataset. Table \ref{table:EventPropertyDistribution} shows event properties classification results. The pink-coloured cells show the model's F1-score when trained on gold-standard dev dataset, F1-scores for minority classes are rather low. As a strategy to overcome class imbalance, we manually over-sample the minority classes for data augmentation and introduce them into the dataset. To this end, we carried out data augmentation through (i) trigger word replacement and (ii) event argument replacement). 

%\begin{figure}[h]
%\centering
%    \includegraphics[width=0.45\textwidth]{image_files/class_imbalance.png}
%    \vspace{-0.7em}
%    \caption{This bar chart shows the `before-and-after data augmentation' class distribution for Polarity, Modality and Intensity. }
%    \label{fig:class_imbalance}
%\end{figure}
%\vspace{-0.7em}

\paragraph{Trigger word replacement} :
FrameNet\footnote{\href{https://framenet.icsi.berkeley.edu}{https://framenet.icsi.berkeley.edu}} was utilized to augment available data and to generate both diverse and valid examples. %FrameNet shares with ACE/ERE a goal of capturing information about events and relations in text.
Authors in \cite{aguilar2014comparison} pointed out that all events, relations, and attributes that were represented by ACE/ERE can be mapped to FrameNet representations through some adjustments. In the selected sentences, we replaced the event trigger words with words (known as \textbf{lexical units} in FrameNet) of the same frame in FrameNet. The idea is to replace the existing trigger word with another valid trigger word while maintaining the same semantic meaning (in FrameNet's term - maintaining the same frame). Through this exercise, we also introduced richer lexical variance in the dataset.:
\vspace{-0.5em}
\begin{exe}
    \item The benchmark for oil prices \textbf{advanced} 1.29\% to \$74.71. \\ Candidates: [\textbf{surged}, \textbf{rose}, \textbf{appreciated}, \textbf{climbed}]
\end{exe}
\vspace{-0.5em}

%\begin{table}[h!]   
%    \centering \small
%    \caption{The mapping between ACE schemas and FrameNet frames}
%    \begin{tabular}{ p{3.5cm} | p{3.5cm} }  \hline
%    \textbf{ACE Schema} & \textbf{FrameNet Frame} \\ \hline
%    Event Trigger & Lexical Unit \\ \hline
%    Event Arguments & Frame Elements \\ \hline
%    Argument Role & Name of Frame Elements \\ \hline
%    Event Type & Frame \\ \hline
%    Entity & Entity \\ \hline
%    \end{tabular}
%    \label{table:mappingACE}
%    \vspace{-0.7em}
%\end{table}

\paragraph{Event argument replacement}
Event argument replacement candidates were chosen from a pool of candidates of the same entity type and the same argument role within the pool of existing annotations, as illustrated below:
\vspace{-0.5em}
\begin{exe}
    \item .....after civil-unrest in \textbf{Libya}...\\
    Candidates: [\textbf{Iraq}, \textbf{Nigeria}, \textbf{Ukraine}]
\end{exe}
\vspace{-0.5em}

After adding augmented data into the training, the green-coloured cells in Table \ref{table:EventPropertyDistribution} show improved F1 scores for minority classes across all three event properties. We add augmented data into the gold-standard dev set to form the \textbf{new development set} and use it to train baseline models for human-in-the-loop active learning.

\subsection{Human-in-the-loop Active Learning} \label{sec:activeLearning}
\begin{figure*}[h]
\centering
    \includegraphics[width=0.85\textwidth]{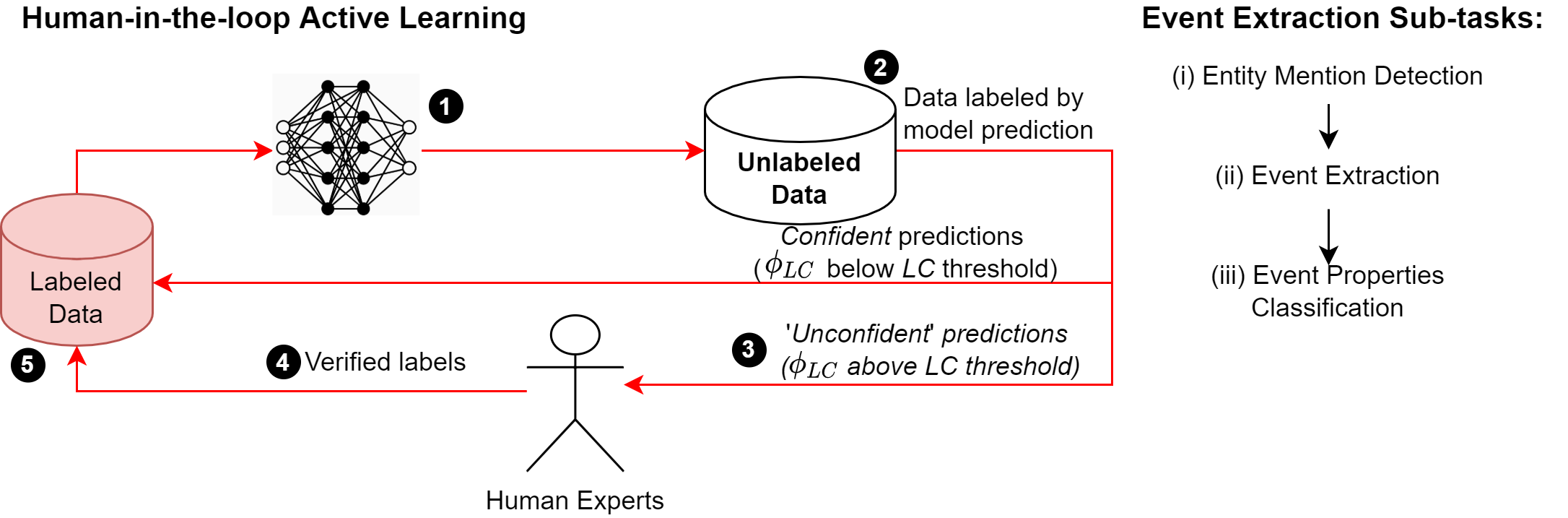}
    \vspace{-0.7em}
    \caption{Human-in-the-loop active learning cycle: it involves (1) training the model with labeled data, (2) using the model to label new data via model prediction, (3) generating sample instances via uncertainty sampling, (4) validating these sample instance by human experts (relabeling if necessary), and (5) adding checked instances to the pool of training data and re-train the models. Steps 1 - 5 are repeated for each event extraction sub-task.}
    \label{fig:active_learning}
\end{figure*}

%\subsection{Bootstrapping / Active Learning} 
%Authors in \citep{abney2002bootstrapping} introduced the concept of \textit{Bootstrapping}. The basic idea is to first train a classifier with a small set of labeled data to classify new unlabeled data. Besides classified labels, a classifier also outputs classification confidence for new data. Then the new data with very high confidence can be included to the training data for next round model training. In \citep{liao2011using}, the authors introduced the idea of \textit{Active Learning} pseudo co-testing, to employ minimum manual manpower to only annotate those low confidence new data. 

%Apart from Supervised Learning approach, the previous work listed above were successful in utilizing transfer learning and active learning in improving the performance event extraction models. Inspired by this, we propose a solution exploring the same approaches to further improve event extraction model despite the limited amount of labeled data.

Active learning is well-motivated in many modern machine learning problems where data may be abundant but labels are scarce or expensive to acquire \cite{settles2009active}. %It allows machine learning classifiers to achieve higher accuracies with fewer training instances by enabling the classifier to interactively query data points. 
\textit{Human-in-the-loop Active Learning} is a strategy of utilizing human expertise in data annotation in a more efficient manner. It is a process of training a model with available labeled data and then uses the model to predict on unlabeled data. Predictions that are `uncertain' (or of low confidence) is then given to human experts for verification. Verified labels are then added into the pool of labeled dataset for training. These predictions are chosen based on \textit{uncertainty sampling}, a sampling strategy to filter out predictions that the model is least confident in. This way we narrow down the scope and have human experts work specifically on these instances. Rather than blindly adding more training data incurring more cost and time, here we target instances that are near the model's decision boundary, they are valuable when labeled correctly and added into the training data to improve model performance.  The whole active learning process is shown in Figure \ref{fig:active_learning}.

\paragraph{\textit{Least Confidence} score}: \textit{Least confidence} score, $\phi_{LC}$ captures how un-confident (or uncertain) a model prediction is. For a probability distribution over a set of labels $y$ for the input $x$, the least confidence score is given by the following equation, where $y^\ast$ is the highest confidence softmax score:
\begin{equation}
\phi_{LC}(x) = (1 - P(y^\ast|x)) \times \frac{n}{n-1}
    \label{eq:confidence_sampling}
\end{equation}
The equation produces a \textit{Least Confidence (LC)} scores to a 0-1 range, where 1 is the most uncertain score while 0 is the most confidence score, $n$ is the number of classes for $y$. The score is normalized for $n$ number of classes by multiplying the result by number of classes, and divided by $n - 1$. Hence it can be used in binary classification as well as multi-class classification . Any model predictions with $\phi_{LC}$ score above the threshold is sampled as they are most likely to be classified wrongly and need to be relabeled by a human annotator.

\subsubsection{Baseline models} \label{sec:baseline}
As baseline for the first round of Active Learning, we trained a number of basic or `vanilla' machine learning models, one for each sub-tasks using the \textbf{new development set} (described in Section \ref{sec:data_augmentation}) as training data and ADJ set as test data (See Table \ref{table:Statistics} for key statistics). These ``vanilla'' models also act as the pilot study demonstrating the use of this dataset in event extraction. The following section describes how these models are trained. 

\paragraph{Entity Mention Detection Model}:
We formalize Entity Mention Detection task as a multi-class token classification. Similar to the approach used in \cite{nguyen2016joint}, we employ BIO annotation schema to assign entity type labels to each token in the sentences. For the model architecture, we use Huggingface's \texttt{BERTForTokenClassification} to fine-tune on this task.

%https://github.com/anoperson/jointEE-NN.

\paragraph{Event Extraction Model}: %Similar to Entity Mention Detection, we formalize the event detection problem as a multi-class classification problem just like in \cite{chen2015event}, \cite{liu2017improving}, and \cite{liu-etal-2018-jointly}. 
We jointly train Event Detection together with Argument Role Prediction using JMEE (Joint Multiple Event Extraction), an event extraction solution proposed by \cite{liu-etal-2018-jointly}. The original version of JMEE uses GloVe word embedding, for this work we used a modified version of JMEE that replaces GloVe with BERT~\cite{devlin-etal-2019-bert} contextualized word embeddings, codes are available    \href{https://github.com/nlpcl-lab/bert-event-extraction}{here}.

\paragraph{Event Properties Classification}:
We use \texttt{BERTForSequenceClassification} model to fine-tune on this task. For every event identified in earlier model, we extract the event `scope' as input for the training. This `scope' is made up of the trigger word(s) being the anchor plus $n$ tokens surrounding it. For the training, we use $n$ = 8. Using the example sentence presented in Figure \ref{fig:annotation}, the `scope' for the second event is ``\textit{oversupply crude oil prices \textbf{plunged} more than 50\% on}''. This sequence of text is fed into the model for event property classification.

\subsubsection{Experiments \& Analysis}
\paragraph{\textit{Least Confidence (LC)} threshold}:
In order to find the optimum sample size for human relabeling, we need to determine the suitable \textit{LC} threshold. We design the uncertainty sampling exercise as a Binary Classification task with two outcomes: \textit{sampled} and \textit{not-sampled}. We experimented with different threshold values to find the optimum sample size for human validation. Apart from being used in the IAA study, the adjudicated (ADJ) set is also used here as the hold-out set to determine the best \textit{LC} threshold. We checked the \textit{sampled} and \textit{not-sampled} instances against the ground-truth in ADJ, and were able to construct the confusion matrix and obtain \textit{Precision}, \textit{Recall} and \textit{F1} scores. Ideally, we want a high \textit{Recall} score (sample as many erroneous cases as possible for human relabeling) and a high \textit{Precision} score as well (identify only relevant instances for correction by keeping correct ones away from being sampled). We experimented with different \textit{LC} threshold value ranging from 0 to 1 in order to find the best threshold that produces \textit{sampled} and \textit{not-sampled} split with the best F1 score (the highest precision-recall pair). %, as shown in Figure \ref{fig:LC_threshold}.
We carry out all iterations of active learning (described next) using the following \textit{LC} thresholds: Entity Mention Detection - 0.60, Trigger Detection - 0.55, Argument Roles Prediction - 0.50, Event Polarity - 0.40, Modality - 0.30, and Intensity - 0.45.

%\begin{itemize}
%    \item True Positive: sampled instances and are erroneous, requiring human relabeling;
%    \item False Positive: sampled instances but are correct;
%    \item False Negative: missed out instances but are erroneous, requiring human relabeling; 
%    \item True Negative: missed out instances and are correct.
%\end{itemize}

%We want to avoid the situation where if threshold value is set too high (closer to 1), then only extremely ``unconfident'' predictions are sampled, potentially resulting in many moderately ``unconfident'' ones being missed. Likewise, we do not want the threshold being set too low (closer to 0) where almost everything is sampled, thus defeating the purpose of Active Learning. 
\paragraph{Experiments}:

\begin{figure*}[htbp]
\centering
    \includegraphics[width=0.95\textwidth]{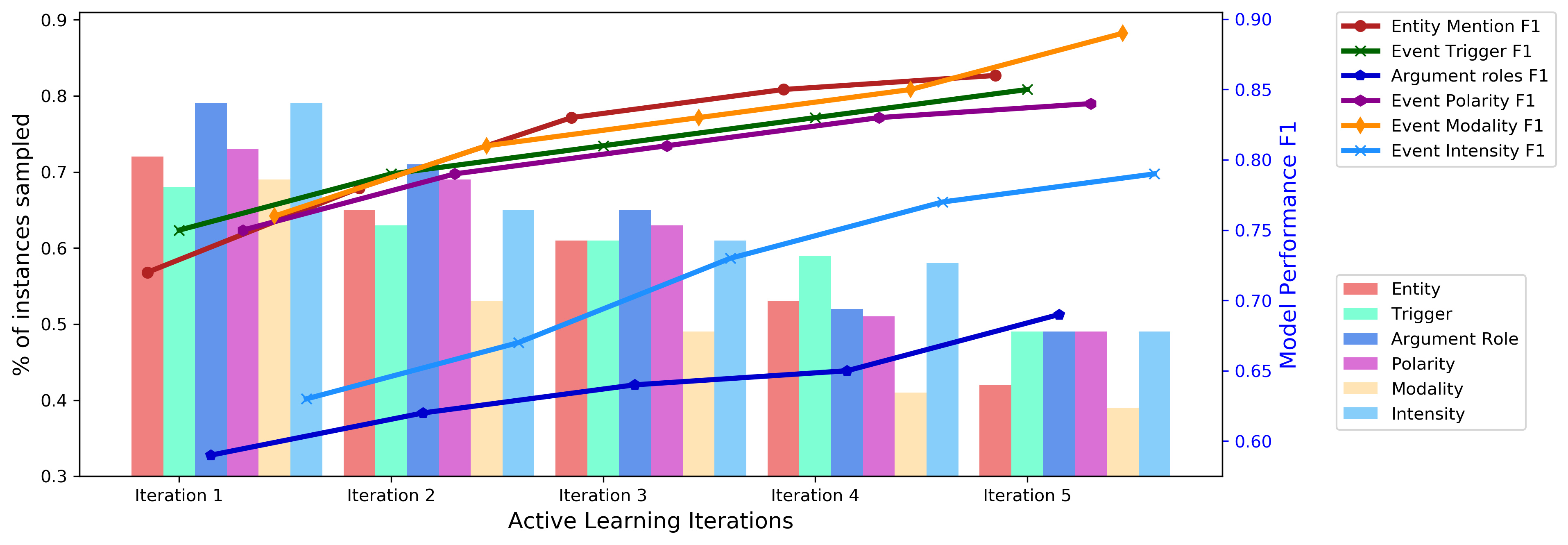}
    \vspace{-0.7em}
    \caption{Results of Active Learning of 5 iterations of Human-in-the-loop Active Learning: (i) the bar chart captures the percentage of data sampled as part of uncertainty sampling; (ii) the line graph shows the model performance (Micro F1 measure) for each sub-tasks. There is an inverse relationship between model performance and percentage of data sampled through uncertainty sampling. See Tables \ref{table:confidence_score} and \ref{table:final_results} for results in tabular form.}
    \vspace{-0.7em}
    \label{fig:AL_results_1}
\end{figure*}

We carried out 5 iterations of active learning, each iteration involves 50 unlabeled crude oil news being labeled through model prediction. Then we ran uncertainty sampling and arranged for two annotators to validate the samples and relabel them if needed. For sentence not sampled, they are deemed 'confident' and therefore being validated/checked by just a single annotator. 
%For sentences with multiple entity mentions and multiple events, if one token among the sentence is sampled, the entire sentence is extracted for human validation. 

\paragraph{Analysis}:
Overall we see improvements in model performance across all sub-tasks. As shown in Figure \ref{fig:AL_results_1}, models performance progressively improved after each iteration. This is because as more annotated training data are added to the training, the more ``confident'' the model gets the fewer instances are sampled under \textit{uncertainty sampling} in each iterations. This inverse relationship is shown in Figure \ref{fig:AL_results_1}. It is clear that as model performance (Micro F1 measure) improves, the percentage of sampled data decreases.

%We noted down the number of instances identified through \textit{uncertainty sampling} where they \textit{least confidence} score, $\phi_{LC}$ above the predefined threshold of 0.3 in Table \ref{table:confidence_score}. At every iteration of model we see better performance but with varying rate of improvement. Results are shown in Figure \ref{fig:AL_results}. The final results after 5 rounds of iteration is captured in Table \ref{table:final_results}.

The least confidence sampling approach is very effective in identifying data points that are near the model's decision boundary. In the case of event type, typically these are events types that can easily confused with other types. For example, the model erroneously classify \textbf{trade tension} as {\fontfamily{qcr}\selectfont Geopolitical-Tension} when the right class should be {\fontfamily{qcr}\selectfont Trade-Tensions}. As the word `tension' exist in both event types, it is understandable why the model makes such a mistake. Least confidence sampling is also able to pick up instances of minority classes. Due to the fact that for minority classes, the model has significantly fewer data to learn from, leading the model to generate predictions that are less `confident'. 

\section{Corpus Statistics and Analysis}
    In total, we managed to produce a final dataset consisting of \textbf{425 documents}, which consist of \textbf{7,059 sentences}, \textbf{10,578 events}, \textbf{22,267 arguments}. The breakdown is shown in Table \ref{table:Statistics}. 
\vspace{-0.7em}
\begin{table}[h!]   
    \centering \small
    \caption{Statistics}
    \begin{tabular}{ l | c | c | c | c }  \hline
    & \multicolumn{2}{c|}{\textbf{Gold-standard}} & \textbf{Aug} & \textbf{5-Iter} \\ \hline
    & \textbf{Dev} & \textbf{Test/ADJ} & \textbf{}& \textbf{Active L.} \\ \hline
    \# documents & 150 & 25 & - & 250  \\ \hline
    \# sentences & 2,557 & 377 & 372& 3,753 \\ \hline
    \# tokens & 68,219 & 9,754 & 12,695 & 99,884 \\ \hline
    \# Entities & 7,120 & 1,970 & 1,838 & 19,417 \\ \hline
    \# Events & 2,943 & 577& 1,061 & 5,997 \\ \hline
    \# Arguments & 5,716 & 1,276 & 1,693 & 13,582 \\ \hline
    \end{tabular}
    \vspace{-0.7em}
    \label{table:Statistics}
\end{table}

%\subsection{Analysis}
%Our event taxonomy is different from SENTiVENT's event typology in \cite{jacobs2021sentivent}. In SENTiVENT the authors defined Event Type and their sub-types and their event triggers are made up of single, continuous and discontinuous token-spans. Exam of discontinuous spans are ``sales....stall'' and ``revenue....rise''. Event types are based on major financial reporting metrics such as ``Profit/loss/Earnings'', ``Revenue'', ``Expense'', ``Sales Volume'' and subtypes such as  ``Increase'', ``Decrease'', ``Stable'', ``Payment'' and etc. In our event typology, on the other hand, define event trigger closer to the definition \textit{Lexical Units} and event arguments as \textit{Frame Elements} in FrameNet. The difference is explained using an example below: ...revenues for 2018, however, are expected to rise 2.1\%.

%\begin{table}[h!]   
%    \centering \small
%    \caption{An example showing the difference between event schema of SENTiVENT and CrudeOilNews (ours)}
%    \begin{tabular}{ p{0.08\textwidth} | p{0.15\textwidth} | p{0.17\textwidth} }  \hline
%    & \textbf{SENTiVENT} & \textbf{CrudeOilNews} \\ \hline
%    Type & Revenue & {\fontfamily{qcr}\selectfont MOVEMENT-UP-GAIN} \\ \hline
%    Subtype & Increase & - \\ \hline
%    Arguments & IncreaseAmount: 2.1\% & Attribute: revenue Difference: 2.1\% \\ \hline
%    \end{tabular}
%    \label{table:difference}
%    \vspace{-0.7em}
%\end{table}

\subsection{Key Characteristics}
We observe a few key characteristics of this corpus that are distinct from ACE2005 and ERE datasets. These need to be taken into consideration when adapting existing event extraction systems or building a new one for this corpus:
\begin{enumerate}
    %\item Majority of sentences contain multiple events, making event extraction more challenging than one-event-one-sentence cases;
    \item Obvious class imbalance in event properties distribution where the majority class outnumbers the minority classes by a large margin (see Table \ref{table:EventPropertyDistribution}). We have attempted to minimize this margin by oversampling minority classes through data augmentation but the margin is still quite substantial;
    \item Homogenous entity types but play different argument roles (e.g., price - non-distinguishable from entity type {\fontfamily{qcr}\selectfont MONEY} or \texttt{UNIT-PRICE}, play different role such as opening price, closing price, and price difference).
    \item Number intensity: Numbers (e.g., price, difference, percentage of change) and dates (including date of the opening price, dates of closing price) are abundant.
\end{enumerate}

\section{Conclusion and Future Work}
Event extraction in the domain of finance and economics at the moment are limited to only company-related events. To contribute to the building of resources in this domain, we have presented CrudeOilNews corpus, a ACE/ERE-like corpus. This corpus contains 425 documents, with around 11,000 events annotated. We have also shared methodologies of how these information were annotated. Inter-annotator agreement is generally substantial and annotator performance is adequate, indicating that the annotation scheme produces consistent event annotations of high quality. 

There are a number of avenues for future work. The main area that can be further explored is to expand the annotation scope to cover more event types. Next, this work can also be expanded to cover event co-reference and event-event relations such as causal-relation, main-sub-event, event-sequence, and contradictory event relation. Besides that, the current sentence-level annotation can be extended to cater for event relations spanning multiple sentences, so that event extraction and relation extract can be done at the document level.

%\begin{table}[!h]
%\begin{center}
%\begin{tabularx}{\columnwidth}{|l|X|}
%      \hline
%      Level&Tools\\
%      \hline
%      Morphology & Pitrat Analyser\\
%      \hline
%      Syntax & LFG Analyser (C-Structure)\\
%      \hline
%     Semantics & LFG F-Structures + Sowa's\\
%     & Conceptual Graphs\\
%      \hline
%\end{tabularx}
%\caption{The caption of the table}
% \end{center}
%\end{table}

% \nocite{*}
\section{Bibliographical References}\label{reference}
%\label{main:ref}

\bibliographystyle{lrec2022-bib}
\bibliography{lrec2022-example}
\bibliographystylelanguageresource{lrec2022-bib}
%\bibliographylanguageresource{languageresource}

\clearpage
\onecolumn
\appendix
\section{Detailed Corpus Statistics}
\begin{table*}[h!]   
    \centering \small
    \caption{Event type distribution and sentence level counts}
    \begin{tabular}{ l |r | c | c | c | c | c}  \hline
    & \multicolumn{2}{c|}{\textbf{Gold Annotation}} & \textbf{Augmented} & \textbf{5-Iter AL} & \multicolumn{2}{c}{\textbf{Final Count}}\\ \hline
    \textbf{Event type} & \textbf{Dev} & \textbf{Test/ADJ} & \textbf{} & \textbf{} & \textbf{\# Instance} & \textbf{Ratio}\\ \hline
    1. Cause-movement-down-loss & 359 & 37 & 179 & 426 & 1,001 & 9.46\% \\ \hline
    2. Cause-movement-up-gain & 72 & 7 & 16 & 83& 178 & 1.68\% \\ \hline
    3. Civil-unrest & 57 & 3 & 43 & 47 & 150 & 1.42\%\\ \hline
    4. Crisis & 19 & 4 & 11 & 34 & 68 & 0.64\%\\ \hline
    5. Embargo & 115 & 7& 44 & 44 & 210 & 1.99\%\\ \hline
    6. Geopolitical-tension & 42 & 10& 25 & 125& 202 & 1.91\%\\ \hline
    7. Grow-strong & 167 & 16& 71 & 280 & 534 & 5.05\%\\ \hline
    8. Movement-down-loss & 697 & 149 & 213 & 1,881 & 2,940 & 27.70\%\\ \hline
    9. Movement-flat & 47 & 2 & 14 & 42 & 105 & 0.99\% \\ \hline
    10. Movement-up-gain & 683 & 178 & 203 & 1,637 & 2,701 & 25.53\% \\ \hline
    11. Negative-sentiment & 116 & 43 & 73 & 307 & 539 & 0.99\%\\ \hline
    12. Oversupply & 65 & 9 & 45 & 112 & 231 & 2.18\% \\ \hline
    13. Position-high & 132 & 33 & 19 & 377 & 561 & 5.03\%\\ \hline
    14. Position-low & 99 & 47 & 24 & 323 & 493 & 4.66\%\\ \hline
    15. Prohibition & 39 & 1 & 3 & 6 & 49 & 0.46\%\\ \hline
    16. Shortage & 31 & 1 & 10 & 5 & 47 & 0.44\%\\ \hline
    17. Slow-weak & 164 & 27 & 52 & 262 & 505 & 4.77\%\\ \hline
    18. Trade-tensions & 39 & 3 & 16 & 6 & 64 & 0.61\%\\ \hline \hline
    \textbf{Total} &\textbf{2,943} & \textbf{577} & \textbf{1,061} & \textbf{5,997} & \textbf{10,578} & \\ \hline
    \end{tabular}
    \label{table:AppDetailDistribution}
    \vspace{-0.7em}
\end{table*}

%\begin{figure}[h]
%\centering
%    \includegraphics[width=0.4\textwidth]{image_files/LC_threshold.png}
%    \vspace{-0.7em}
%    \caption{Determining the best \textit{LC} threshold that results in the best F1-Score (the highest precision-recall pair).}
%    \label{fig:LC_threshold}
%\end{figure}

\section{Active Learning details}
\begin{table*}[h!]   
    \centering \small
    \caption{The percentage of instances (not number of sentences) sampled through uncertainty sampling ($\phi_{LC}$ score above the threshold value). In each active learning iteration, 50 unlabeled crude oil news were randomly selected  and labeled through model prediction. See Figure \ref{fig:AL_results_1} for results in graph form.}
    \begin{tabular}{ |l | c|  c|  c | c |  c | c |}  \hline
     & \textbf{Entity} & \textbf{Trigger} & \textbf{Arguments} & \textbf{Polarity} & \textbf{Modality} & \textbf{Intensity} \\ \hline
    \textbf{Threshold} & 0.6 & 0.55 & 0.50 & 0.40 & 0.30 & 0.45 \\ \hline \hline
    \textbf{Iter.} & \% of \# tokens & \% of \# tokens & \% of Trigger-Entity Pair & \% of events & \% of events & \% of events \\ \hline
    1 &  72 & 68 & 75 & 73 & 69 & 79 \\ \hline
    2 &  65 & 63 & 71 & 69 & 53 & 65 \\ \hline
    3 &  61 & 61 & 65 & 63 & 49 & 61 \\ \hline
    4 &  53 & 59 & 62 & 51 & 41 & 58 \\ \hline
    5 &  42 & 49 & 51 & 49 & 39 & 49 \\ \hline
    \end{tabular}
    \label{table:confidence_score}
\end{table*}

%avg_bar1 = (0.72,0.65,0.61,0.53,0.42)  Entity
%avg_bar2 = (0.68,0.63,0.61,0.59,0.49) Trigger
%avg_bar3 = (0.75,0.71,0.65,0.62, 0.51) Argument roles
%avg_bar4 = (0.53, 0.49,0.43,0.41,0.39) Polarity
%avg_bar5 = (0.49,0.43,0.42,0.41,0.35) Modality
%avg_bar6 = (0.69,0.65,0.61,0.58,0.49) Intensity

 \begin{table*}[h!]   
    \centering \small
        \caption{Model performance (Micro F1-score) across varying amount of training data. As the amount of training data increases, the performance of each model increases as well. System evaluation is done on Gold-standard Test/ADJ Set. See Figure \ref{fig:AL_results_1} for results in graph form.}
    \begin{tabular}{ | l |l |c | c | c | c | c| c |}  \hline
     \textbf{Iter.} & \textbf{Training Set} & \textbf{Entity} & \textbf{Trigger} & \textbf{Argument} &  \textbf{Polarity} & \textbf{Modality} & \textbf{Intensity} \\ \hline 
    - & \small{Gold Dev} & 
        \small{0.71} & \small{0.74} & \small{0.56} & 
        \small{0.74} & \small{0.71} & \small{0.75} \\ \hline
    - & \small{Gold Dev + Augmented (New Dev)} & 
        \small{0.72} & \small{0.75} & \small{0.57} & 
        \small{0.75} & \small{0.73} & \small{0.75} \\ \hline
    1 & \small{New Dev + 50 docs} & 
        \small{0.72} & \small{0.75} & \small{0.59} & 
        \small{0.75} & \small{0.76} & \small{0.73} \\ \hline   
    2 & \small{New Dev + 100 docs} & 
        \small{0.78} & \small{0.79} & \small{0.62} & 
        \small{0.79} & \small{0.81} & \small{0.77} \\ \hline   
    3 & \small{New Dev + 150 docs} & 
        \small{0.83} & \small{0.81} & \small{0.64} & 
        \small{0.81} & \small{0.83} & \small{0.81} \\ \hline    
    4 & \small{New Dev + 200 docs} & 
        \small{0.85} & \small{0.83}& \small{0.65} & 
        \small{0.83} & \small{0.85} & \small{0.82} \\ \hline
    5 & \small{New Dev + 250 docs} & 
        \small{0.86} & \small{0.85} & \small{0.69} & 
        \small{0.84} & \small{0.89} & \small{0.83} \\ \hline   
      \end{tabular}
    \vspace{-0.7em}
    \label{table:final_results}
\end{table*}

Note: \textbf{New development set} the baseline model described in Section \ref{sec:baseline}.

\clearpage
\section{Entity Mention Types} \label{app:entity}
 \begin{table}[h!]   
    \begin{center}
    \caption{List of Entity Types}
    \begin{tabular}{ l | p{0.7\linewidth}}  \hline
    \textbf{Entity Type} & \textbf{Examples} \\ \hline
    1. Commodity & \textit{oil, crude oil, Brent, West Texas Intermediate (WTI), fuel, U.S Shale, light sweet crude, natural gas} \\ \hline
    2. Country** & \textit{Libya, China, U.S, Venezuela, Greece} \\ \hline
    3. Date** & \textit{1998, Wednesday, Jan. 30, the final quarter of 1991, the end of this year}  \\ \hline
    4. Duration** & \textit{two years, three-week, 5-1/2-year, multiyear, another six months} \\ \hline
    5. Economic Item & \textit{economy, economic growth, market, economic outlook, employment data, currency, commodity-oil}  \\ \hline
    6. Financial attribute & \textit{supply, demand, output, production, price, import, export} \\ \hline
    7. Forecast target & \textit{forecast, target, estimate, projection, bets} \\ \hline
    8. Group & \textit{global producers, oil producers, hedge funds, non-OECD, Gulf oil producers} \\ \hline
    9. Location** & \textit{global, world, domestic, Middle East, Europe} \\ \hline
    10. Money** & \textit{\$60, USD 50}  \\ \hline
    11. Nationality** & \textit{Chinese, Russian, European, African} \\ \hline
    12. Number** & (any numerical value that does not have a currency sign) \\ \hline
    13. Organization** & \textit{OPEC, Organization of Petroleum Exporting Countries, European Union, U.S. Energy Information Administration, EIA} \\ \hline
    14. Other activities & (free text) \\ \hline
    15. Percent** & \textit{25\%, 1.4 percent} \\ \hline
    16. Person** & \textit{Trump, Putin} (and other political figures) \\ \hline
    17. Phenomenon & (free text) \\ \hline
    18. Price unit & \textit{\$100-a-barrel, \$40 per barrel, USD58 per barrel}  \\ \hline
    19. Production Unit & \textit{170,000 bpd, 400,000 barrels per day, 29 million barrels per day} \\ \hline
    20. Quantity & \textit{1.3500 million barrels, 1.8 million gallons, 18 million tonnes} \\ \hline
    21. State or province** & \textit{Washington, Moscow, Cushing, North America} \\ \hline
    \end{tabular}
    \vspace{-1.0em}
    \label{table:EntityTypes}
    \end{center}
\end{table}

\clearpage
\section{Event Schema} \label{app:schema}

%%%%%%%%%%% Movement-down-loss, movement-up-gain, movement-flat %%%%%%%
\subsection{Movement-down-loss, Movement-up-gain, Movement-flat}
\textbf{Example sentence: } [Globally] [crude oil] [futures] \textbf{surged} [\$2.50] to [\$59 per barrel] on [Tuesday].
 \begin{table}[h]   
    \centering
    %\caption{table:Event Arguments for \textbf{Movement-down-loss}, \textbf{Movement-up-gain}, and \textbf{Movement-flat}}
    \begin{tabular}{ | p{0.2\linewidth} | p{0.6\linewidth} | p{0.15\linewidth} |}  \hline
    \textbf{Role} & \textbf{Entity Type} & \textbf{Argument Text}\\ \hline
    
     Type & Nationality, Location & globally \\ \hline
     Place & Country, Group, Organization, Location, State or province, Nationality & \\ \hline
     Supplier\_consumer & Organization, Country, State\_or\_province, Group, Location & \\ \hline
     Reference\_point\_time & Date & Tuesday \\ \hline
     Initial\_reference\_point & Date &  \\ \hline
     Final\_value & Percentage, Number, Money, Price\_unit, Production\_unit, Quantity & \$59 per barrel \\ \hline
     Initial\_value & Percentage, Number, Money, Price\_unit, Production\_unit, Quantity &  \\ \hline
     Item & Commodity, Economic\_item & crude oil \\ \hline
     Attribute & Financial\_attribute & futures\\ \hline
     Difference & Percentage, Number, Money, Production\_unit, Quantity & \$2.50 \\ \hline
     Forecast & Forecast\_target &  \\ \hline
     Duration & Duration & \\ \hline
     Forecaster & Organization & \\ \hline 
    \end{tabular}
    \vspace{-1.0em}
\end{table}

%%%%%% Caused-movement-downloss, caused-movement-up-gain %%%%%%%%%%
\subsection{Caused-movement-down-loss, Caused-movement-up-gain}
\textbf{Example sentence:} The [IMF] earlier said it \textbf{reduced} its [2018] [global] [economic growth] [forecast] to [3.30\%] from a [July] forecast of [4.10\%].
 \begin{table}[h]
    \centering
    \begin{tabular}{ | p{0.2\linewidth} | p{0.6\linewidth} | p{0.15\linewidth} |} \hline
    \textbf{Role} & \textbf{Entity Type} & \textbf{Argument Text}\\ \hline
    
     Type & Nationality, Location & global \\ \hline
     Place & Country, Group, Organization, Location, State or province, Nationality & West African, European \\ \hline
     Supplier\_consumer & Organization, Country, State\_or\_province, Group, Location & \\ \hline
     Reference\_point\_time & Date & 2018 \\ \hline
     Initial\_reference\_point & Date & July \\ \hline
     Final\_value & Percentage, Number, Money, Price\_unit, Production\_unit, Quantity & 3.30\%\\ \hline
     Initial\_value & Percentage, Number, Money, Price\_unit, Production\_unit, Quantity & 4.10\% \\ \hline
     Item & Commodity, Economic\_item & economic growth \\ \hline
     Attribute & Financial\_attribute & \\ \hline
     Difference & Percentage, Number, Money, Production\_unit, Quantity & \\ \hline
     Forecast & Forecast\_target & forecast \\ \hline
     Duration & Duration & \\ \hline
     Forecaster & Organization & IMF \\ \hline 
    \end{tabular}
    \vspace{-1.0em}
\end{table}

\subsection{Position-high, Position-low}
\textbf{Example sentence:} The IEA estimates that U.S. crude oil is expected to seek higher ground until reaching a [5-year] \textbf{peak} in [late April] of about [17 million bpd].
 \begin{table}[ht]   
    \centering
    \begin{tabular}{ | p{0.2\linewidth} | p{0.6\linewidth} | p{0.15\linewidth} |}  \hline
    \textbf{Role} & \textbf{Entity Type} & \textbf{Argument Text}\\ \hline
         Reference\_point\_time & Date &  late April \\ \hline
     Initial\_reference\_point & Date &  \\ \hline
     Final\_value & Percentage, Number, Money, Price\_unit, Production\_unit, Quantity &  17 million bpd\\ \hline
     Initial\_value & Percentage, Number, Money, Price\_unit, Production\_unit, Quantity &  \\ \hline
     Item & Commodity, Economic\_item & \\ \hline
     Attribute & Financial\_attribute & \\ \hline
     Difference & Percentage, Number, Money, Production\_unit, Quantity &  \\ \hline
     Duration & Duration & 5-year \\ \hline
    \end{tabular}
\end{table}

\clearpage
\subsection{Slow-weak, Grow-strong}
\textbf{Example sentence:}  [U.S.] [employment data] \textbf{strengthens} with the euro zone. 
 \begin{table}[h]   
    \centering
    \begin{tabular}{ | p{0.2\linewidth} | p{0.6\linewidth} | p{0.15\linewidth} |} \hline
    \textbf{Role} & \textbf{Entity Type} & \textbf{Argument Text}\\ \hline
    
     Type & Nationality, Location & \\ \hline
     Place & Country, Group, Organization, Location, State or province, Nationality & U.S. \\ \hline
     Supplier\_consumer & Organization, Country, State\_or\_province, Group, Location & \\ \hline
     Reference\_point\_time & Date &  \\ \hline
     Initial\_reference\_point & Date &  \\ \hline
     Final\_value & Percentage, Number, Money, Price\_unit, Production\_unit, Quantity &  \\ \hline
     Initial\_value & Percentage, Number, Money, Price\_unit, Production\_unit, Quantity &  \\ \hline
     Item & Commodity, Economic\_item & employment data \\ \hline
     Attribute & Financial\_attribute & \\ \hline
     Difference & Percentage, Number, Money, Production\_unit, Quantity &  \\ \hline
     Forecast & Forecast\_target &  \\ \hline
     Duration & Duration & \\ \hline
     Forecaster & Organization & \\ \hline 
    \end{tabular}
    \vspace{-1.0em}
\end{table}
    
%%%%%% Prohibiting   %%%%%%%
\subsection{Prohibiting}
\textbf{Example sentence:} [Congress] \textbf{banned} most [U.S.] [crude oil] [exports] on [Friday] after price shocks from the 1973 Arab oil embargo.
 \begin{table}[h]   
    \centering
    \begin{tabular}{ | p{0.2\linewidth} | p{0.6\linewidth} | p{0.15\linewidth} |}  \hline
    \textbf{Role} & \textbf{Entity Type} & \textbf{Argument Text}\\ \hline
     Imposer & Organization, Country, Nationality, State or province, Person, Group, Location &  Congress \\ \hline
     Imposee & Organization, Country, Nationality, State or province, Group & U.S. \\ \hline
     Item & Commodity, Economic\_item & crude oil \\ \hline
     Attribute & Financial\_attribute & exports \\ \hline
     Reference\_point\_time & Date & Friday \\ \hline
     Activity & Other\_activities & \\ \hline
    \end{tabular}
    \vspace{-1.0em}
\end{table}

%%%%%% Oversupply %%%%%%%%%%
\subsection{Oversupply}
\textbf{Example sentence:} [Forecasts] for an [crude] \textbf{oversupply} in [West African] and [European] [markets] [early June] help to push the Brent benchmark down more than 20\% January.
\vspace{-0.5em}
 \begin{table}[h]   
    \centering
    \begin{tabular}{ | p{0.2\linewidth} | p{0.6\linewidth} | p{0.15\linewidth} |} \hline
    \textbf{Role} & \textbf{Entity Type} & \textbf{Argument Text}\\ \hline
     Place & Country, Group, Organization, Location, State or province, Nationality & West African, European \\ \hline
     Reference\_point\_time & Date & this year \\ \hline
     Item & Commodity & crude \\ \hline
     Attribute & Financial\_attribute & markets \\ \hline
     Difference & Production\_unit & \\ \hline
     Forecast & Forecast\_target & forecasts \\ \hline
    \end{tabular}
\end{table}

\clearpage

%%%%% Shortage %%%%%%
\subsection{Shortage}
\textbf{Example Sentence:} Oil reserves are within ``acceptable'' range in most oil consuming countries and there is no \textbf{shortage} in [oil] [supply] [globally], the minister added.
 \begin{table}[h]   
    \centering
    \begin{tabular}{ | p{0.2\linewidth} | p{0.6\linewidth} | p{0.15\linewidth} |} \hline
    \textbf{Role} & \textbf{Entity Type} & \textbf{Argument Text}\\ \hline
     Place & Country, State or province, Location, Nationality &  Congress \\ \hline
     Item & Commodity & crude oil \\ \hline
     Attribute & Financial\_attribute & exports \\ \hline
     Type & Location & globally \\ \hline
     Reference\_point\_time & Date &  \\ \hline
    \end{tabular}
\end{table}

%%%% Civil unrest %%%%%
\subsection{Civil Unrest}
\textbf{Example sentence:} The drop in oil prices to their lowest in two years has caught many observers off guard, coming against a backdrop of the worst \textbf{violence} in [Iraq] [this decade].
 \begin{table}[h]   
    \centering
    \begin{tabular}{ | p{0.2\linewidth} | p{0.6\linewidth} | p{0.15\linewidth} |}  \hline
    \textbf{Role} & \textbf{Entity Type} & \textbf{Argument Text}\\ \hline
     Place & Country, State or province, Location, Nationality & Iraq  \\ \hline
     Reference\_point\_time & Date & this decade \\ \hline
    \end{tabular}
\end{table}

\subsection{Embargo}
\textbf{Example sentence:} The [Trump administration] imposed a ``strong and swift'' economic \textbf{sanctions} on [Venezuela] on [Thursday].
 \begin{table}[h]   
    \centering
    \begin{tabular}{ | p{0.2\linewidth} | p{0.6\linewidth} | p{0.15\linewidth} |}  \hline
    \textbf{Role} & \textbf{Entity Type} & \textbf{Argument Text}\\ \hline
     Imposer & Organization, Country, Nationality, State or province, Person, Group, Location &  Trump administration \\ \hline
     Imposee & Organization, Country, Nationality, State or province, Group & Venezuela \\ \hline
     Reference\_point\_time & Date & Thursday \\ \hline
    \end{tabular}
\end{table}\\
Note: `Imposee' is not formally a word, but used here as a shorter version of ``Party whom the action was imposed on.

\clearpage
%%%%%% Geo-political Tension %%%%%%%%%%
\subsection{Geo-political Tension}
\textbf{Example sentence: } \textbf{Deteriorating relations} between [Iraq] and [Russia] [first half of 2016] ignited new fears of supply restrictions in the market.
\vspace{-0.5em}
 \begin{table}[h]   
    \centering
    \begin{tabular}{ | p{0.2\linewidth} | p{0.6\linewidth} | p{0.15\linewidth} |} \hline
    \textbf{Role} & \textbf{Entity Type} & \textbf{Argument Text}\\ \hline
     Participating\_countries & Country, Group, Organization, Location, State or province, Nationality & U.S., China \\ \hline
     Reference\_point\_time & Date & early June \\ \hline
    \end{tabular}
    \vspace{-1.0em}
\end{table}

%%%%%  Crisis %%%%%%
\subsection{Crisis}
\textbf{Example Sentence:} Asia 's diesel consumption is expected to recover this year at the second weakest level rate since the [2014] [Asian] [financial] \textbf{crisis}.
 \begin{table}[h]   
    \centering
    \begin{tabular}{ | p{0.2\linewidth} | p{0.6\linewidth} | p{0.15\linewidth} |} \hline
    \textbf{Role} & \textbf{Entity Type} & \textbf{Argument Text}\\ \hline
     Place & Country, State or province, Location, Nationality & Asian \\ \hline
     Reference\_point\_time & Date & this year \\ \hline
     Item & Commodity, Economic\_item & financial \\ \hline 
    \end{tabular}
\end{table}

\subsection{Negative Sentiment}
\textbf{Example sentence:} Oil futures have dropped due to \textbf{concern} about softening demand growth and awash in crude.

Note: \textbf{Negative Sentiment} is a special type of event, where majority of the time it contains just the trigger words such as \textit{concerns, worries, fears} and 0 event arguments.

\end{document}